\documentclass[letterpaper, 10 pt, conference]{ieeeconf}  

\IEEEoverridecommandlockouts                              

\usepackage{graphics} 
\usepackage{epsfig} 
\usepackage{amsmath} 
\usepackage{amssymb}  

\DeclareMathOperator{\ave}{ave}

\usepackage[]{subfigure}

\title{\LARGE \bf
RAMP: Reaction-Aware Motion Planning of Multi-Legged Robots for Locomotion in Microgravity

}

\author{Warley F. R. Ribeiro$^{1}$, Kentaro Uno$^{1}$, Masazumi Imai$^{1}$, Koki Murase$^{1}$ and Kazuya Yoshida$^{1}$
    \thanks{This work was not supported by any organization}
    \thanks{$^{1}$Warley F. R. Ribeiro, Kentaro Uno, Masazumi Imai, Koki Murase and Kazuya Yoshida are with the Department of Aerospace Engineering, Graduate School of Engineering, Tohoku University, Sendai, Japan. 
        {\tt\small \{warley, imai.masazumi.p2, murase.koki.q2\}[at]dc.tohoku.ac.jp and \{unoken, yoshida.astro\}[at]tohoku.ac.jp}
    }
}

\begin{document}

\maketitle
\thispagestyle{empty}
\pagestyle{empty}

\begin{abstract}

Robotic mobility in microgravity is necessary to expand human utilization and exploration of outer space. Bio-inspired multi-legged robots are a possible solution for safe and precise locomotion. However, a dynamic motion of a robot in microgravity can lead to failures due to gripper detachment caused by excessive motion reactions. We propose a novel Reaction-Aware Motion Planning (RAMP) to improve locomotion safety in microgravity, decreasing the risk of losing contact with the terrain surface by reducing the robot's momentum change. RAMP minimizes the swing momentum with a Low-Reaction Swing Trajectory (LRST) while distributing this momentum to the whole body, ensuring zero velocity for the supporting grippers and minimizing motion reactions. We verify the proposed approach with dynamic simulations indicating the capability of RAMP to generate a safe motion without detachment of the supporting grippers, resulting in the robot reaching its specified location. We further validate RAMP in experiments with an air-floating system, demonstrating a significant reduction in reaction forces and improved mobility in microgravity.

\end{abstract}

\section{Introduction}

Mobile robots are part of the essential technologies necessary to increase human presence and exploration of extraterrestrial bodies. Space has, however, several extreme circumstances different from the regular mobile robots on Earth, such as drastic temperature variation, high radiation, and low gravity. A more specific condition is microgravity, present in two crucial situations: orbiting space stations and exploration of asteroids.

By exploring Small Solar System Bodies (SSSBs), such as asteroids and comets, we can obtain significant scientific data to understand the process of formation and evolution of our Solar System. Those celestial objects also showed substantial amounts of water and organic matter, which could give us insights into the process of life formation. SSSBs can also threaten our planet if they cross Earth's orbit, and impact mitigation missions need to be a concrete alternative for our safety. And lastly, SSSBs are also rich in valuable materials for use on Earth and further space development~\cite{Badescu2013}. All those scenarios need to rely on mobile robots capable of performing interactive activities on the surface of the SSSB, such as in-situ investigation, terrain mapping, and resource sampling and analysis.

Another use of mobile robots is assisting astronauts in orbiting stations, such as the International Space Station~(ISS). Robots could achieve simple tasks, such as helping with setting up experiments and recording the results of the tests to optimize the valuable crew time~\cite{russell2006applying}. High-capable robots can also perform more complex activities, such as maintaining the components necessary for habitable conditions, transporting and organizing the cargo inside the station and assembling new structures and modules.

\begin{figure}[t]
\centering
\includegraphics[width=0.48\textwidth]{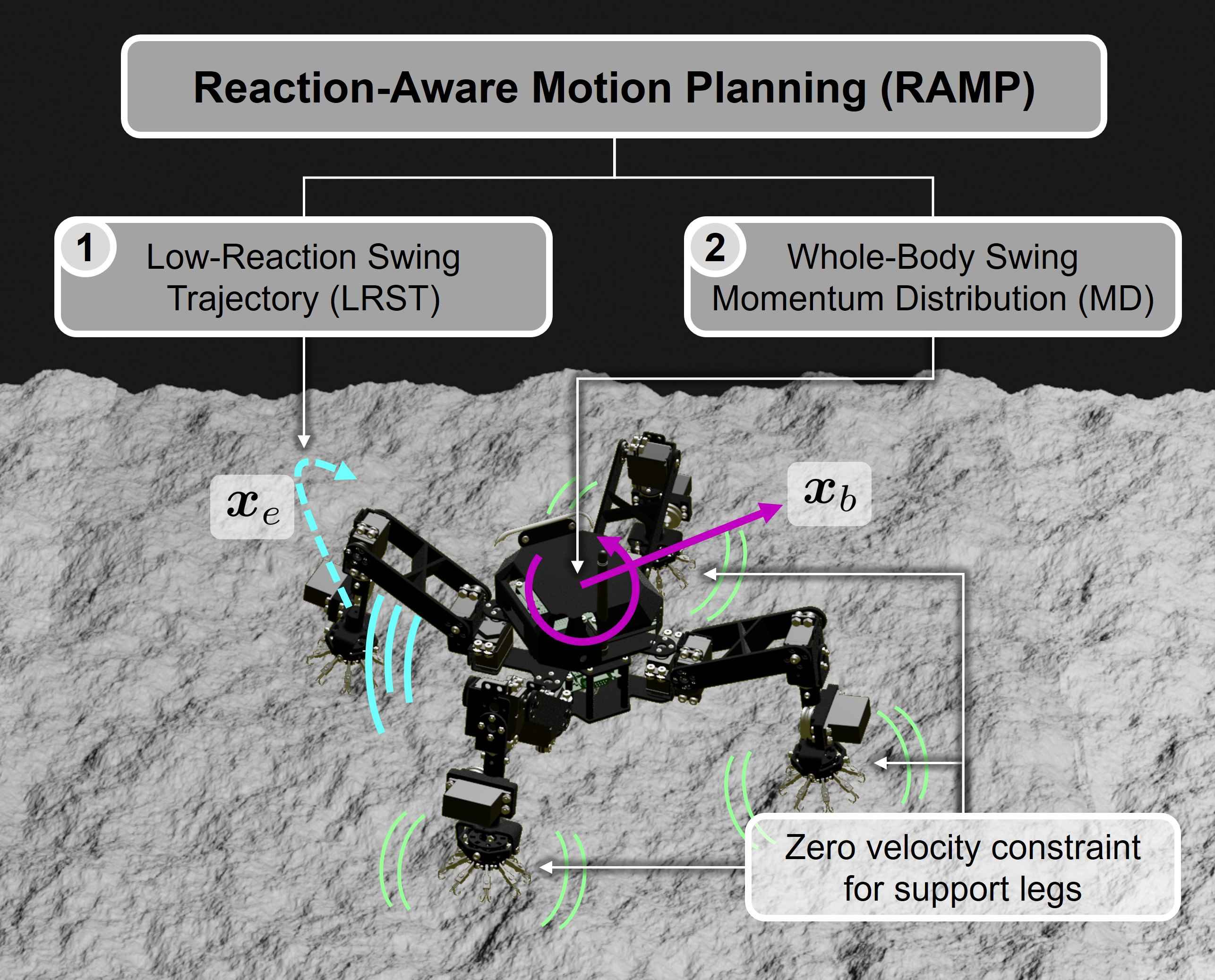}
\caption{Scheme for the Reaction-Aware Motion Planning (RAMP). The trajectory of the swing limb $\boldsymbol{x}_e$ is optimized to minimize motion reactions using a Low-Reaction Swing Trajectory (LRST) algorithm. The motion of the base $\boldsymbol{x}_b$ is generated to balance the swing reactions with the Whole-Body Swing Momentum Distribution (MD) technique. A zero constraint velocity is applied for all supporting limbs to mitigate motion reactions at the contact points. RAMP can mitigate motion reactions to reduce the risk of detachment from the surface.}
\label{fig:RAMP}
\end{figure}

\subsection{Related Works}

The hopping mobility technique is the only robotic surface locomotion method employed in actual missions to SSSBs up to this date. A robot uses an internal actuator, such as a flywheel, to create a striking reaction force from the surface to generate a jump with a ballistic trajectory. Due to the microgravity condition, small and lightweight actuators can produce high hops within the escape velocity of the celestial body~\cite{yoshimitsu1999new}. The twin rovers MINERVA-II-1~\cite{yoshimitsu2022asteroid} and the lander MASCOT~\cite{ho2017mascot} successfully demonstrated this method in the Japanese Hayabusa2 mission to the asteroid Ryugu~\cite{tsuda2020hayabusa2}.
A fourth robot, MINERVA-II-2, was also aboard the Hayabusa2 mission, equipped with four different actuators to produce hops~\cite{nagaoka2015development}. However, the second MINERVA-II rover did not accomplish its mobility objective due to a malfunction in the main computer. 
The development of hopping robots continues to evolve, with more robots incorporating new mobility concepts and strategies~\cite{hockman2017design,kobashi2020tumbling}.

When we consider past and current robots for assistance in a space station, the primary locomotion method is that of the free-flying robots. By utilizing propellant gases or fans to generate the reaction forces, small robots can move inside the tridimensional space of the ISS thanks to the negligible gravity. Astrobee~\cite{bualat2018astrobee}, CIMON~\cite{schroder2018gnc} and Int-Ball~\cite{mitani2019int} are some successful examples of free-flying robots with autonomous mobility and task-performing capabilities operating in the ISS.

Bio-inspired legged robots were proposed as a viable general solution for mobility in microgravity~\cite{yoshida2002novel}. Considering uncontrollable hops inside a space station can be dangerous for the astronauts, and propellant-based free-flying motion is limited for SSSBs without an atmosphere, those robots do not work in any microgravity locomotion context. Also, fast and accurate mobility is challenging with robots based on free-flying and hopping techniques. Multi-legged robots with surface attachment capabilities can achieve precise and stable locomotion in space stations and small celestial bodies, although there is additional complexity regarding fabrication and control.

Robonaut 2 was a humanoid robot with climbing legs that flew to the ISS as a robot to assist astronauts but did not demonstrate mobility in microgravity~\cite{diftler2012robonaut}. As an example of multi-legged robots for microgravity locomotion, we have the LEMUR series developed by NASA JPL, where the latest version can have microspine grippers for rough surfaces of SSSBs, or gecko-inspired adhesive grippers for smooth surfaces in the ISS~\cite{parness2017lemur}. A similar robot with microspines was also proposed, including stochastic grasping modelling~\cite{wen2022stochastic}. Another example is ReachBot, which includes extendable limbs to improve the reachability of a compact designed robot~\cite{schneider2022reachbot}. 

\subsection{Objectives and Contributions}

Besides the complexity of a system with multiple links and developing a reliable grasping mechanism, the safe mobility of multi-legged robots in microgravity is still an issue. Although the lack of gravity facilitates static equilibrium, quasi-static legged mobility is not a suitable solution for space robots, as the time constraints are significantly severe. It is required to perform dynamic evaluations to incorporate the inertial acceleration and forces, as the robot's motion can cause the detachment of the grippers. Chacin et al. proposed a compliant motion control considering the friction forces conditions for equilibrium~\cite{chacin2009motion}. Yuguchi et al. proposed a different approach, using a reactionless method for a bipedal robot to reduce the ground reaction forces~\cite{yuguchi2016verification}. However, precise friction characteristics are not available before exploration missions, complicating the estimation of forces acting on the robot. Moreover, reactionless control was demonstrated only for a single contact point and could cause considerable changes in the robot's pose, inducing kinematic singularities.

In this paper, we propose a novel Reaction-Aware Motion Planning (RAMP) to improve the safety of multi-legged locomotion in microgravity by decreasing the motion reactions. We also aim to provide low-reaction alternatives to the reactionless control to prevent singular configurations. First, we introduce our approach to mitigate the change in the robot's momentum in two different aspects. The first is to minimize the reactions generated from the swing leg motion. And the second is the whole-body momentum distribution to partially or fully compensate for the swing momentum, constraining the velocity of contact points to zero. Later, we verify the increased performance of the proposed motion planning in dynamic simulations of realistic scenarios of multi-legged mobility in microgravity. And finally, we validate the approach in an emulated microgravity facility.

\section{Reaction-Aware Motion Planning}

In a microgravity condition, we can neglect the effects of the gravity force when considering the elements that can cause the detachment of the grippers. Assuming that only the ground reaction forces are acting on the robot, we can limit the detachment analysis to the maximum pulling force the gripper can endure before undesired separation from the surface. And the only source of the pulling forces under this microgravity condition is the robot's self-movement, either due to the swing of the legs to the next grasping position or the main body to reach the desired location. In this study, we define \emph{Motion Reaction} as the pulling or pushing action of the supporting grippers induced by the robot's movement.

In the case of quasi-static walking, the motion reactions are virtually null, and no surface separation should occur. But responsive and agile robots are needed for future space exploration and human assistance, and dynamic locomotion is necessary to achieve our space utilization goals. The proposed Reaction-Aware Motion Planning (RAMP) addresses the safety locomotion issue by generating a swing motion for the leg with the minimum motion reaction. Additionally, the robot performs an additional movement using the remaining degrees of freedom to drive its main body, distributing the generated motion reactions. The motion plan also adds zero velocity constraint for the supporting grippers, assuring that no excessive pulling motion happens. The momentum distribution with zero velocity at the contact points further decreases motion reactions and ground reaction forces that could cause detachment. Fig.~\ref{fig:RAMP} shows an outline of the proposed motion planning.

\subsection{Momentum Variation Approach}

A simple way to quantify the pulling action of the grippers due to motion reactions is by computing the total internal force induced by the planned robot's movement. This force is the derivative of the resultant momentum of the motion generated by the robot. Equation~(\ref{eq:momentum}) describes the momentum $\boldsymbol{\mathcal{L}}$ of an articulated robot with multiple rigid links and $n$ legs under microgravity. $\boldsymbol{H}_{b}$ and $\boldsymbol{H}_{bm,i}$ are the inertia matrices of the base link and the base-manipulator coupling, respectively. The vector with the tridimensional position of the base link is $\boldsymbol{x}_{b}$, and $\boldsymbol{\phi}_i$ is the vector with the angular position of all articulated joints.

\begin{equation}
  \begin{aligned}
    \boldsymbol{\mathcal{L}} = 
    \boldsymbol{H}_{b} \dot{\boldsymbol{x}}_{b} + 
    \sum_{i=1}^{n} \boldsymbol{H}_{bm,i} \dot{\boldsymbol{\phi}}_i 
   \end{aligned} 
   \label{eq:momentum}
\end{equation}

Therefore, by minimizing the variation in the robot's momentum, we can reduce the motion reactions causing the pulling movement of the grippers, which is the primary reason for surface separation under microgravity, increasing the locomotion stability.

\subsection{Low-Reaction Swing Trajectory}

The swing motion of a leg from its current position to the location of the next grasping point needs to be in the feasible space of the robot and should avoid collisions with the environment. We include a third condition to the trajectory generation algorithm: minimizing momentum variation to lower the motion reactions, preventing the gripper detachment in microgravity conditions.

A low-reaction trajectory generation was proposed based on optimizing the coefficients of a polynomial curve to find the trajectory with minimum momentum variation, reaching the desired step height to avoid collisions~\cite{ribeiro2022low}.
Equation~(\ref{eq:opt_bez}) shows the details of the optimization problem that constrains the end-effector trajectory $\boldsymbol{x}_e(t)$ from the initial time $t_0$ to the final time $t_f$ by a 7th-order Bezier curve of coefficients $\boldsymbol{a}_i$. The optimization is also subject to the feasible angular position of joints $\boldsymbol{\phi}$ bounded by the minimum and maximum values, $\boldsymbol{\phi}_{\min}$ and $\boldsymbol{\phi}_{\max}$ respectively.

\begin{equation}
\label{eq:opt_bez}
    \begin{aligned}
        \min_{\boldsymbol{a}_3,\boldsymbol{a}_4} \quad & J_1({\dot{\mathcal{L}}}(t)) + J_2(x_z(t))      \\
        \textrm{s.t.} \quad 
            & \boldsymbol{x}_e(t) = \sum_{i=0}^7  B_i(t) \boldsymbol{a}_i \quad \quad t_0 \leq t \leq t_f\\
            & \boldsymbol{\phi}_{\min} \leq \boldsymbol{\phi}(t) \leq \boldsymbol{\phi}_{\max} \\
    \end{aligned}
\end{equation}

The objective functions $J_1$ and $J_2$ incorporate the momentum variation and the step height $h$ of the desired trajectory, including selectable coefficients $k$ to adjust the weight of each term. 

\setlength{\arraycolsep}{0.1em}
\begin{eqnarray}
    J_1({\dot{\mathcal{L}}}(t))=&&k_1\max({\dot{\mathcal{L}}}(t)) \label{eq:objective1}\\
    J_2(x_z(t))                     =&&k_2\left|h-\max(x_z(t))\right| \nonumber \\
                                     &+&k_3  \left| h - \ave \left(x_z(t)\right) \right|  \label{eq:objective2}
\end{eqnarray}

\subsection{Whole-Body Swing Momentum Distribution}

The Low-Reaction Swing Trajectory (LRST) generation method successfully reduces motion reactions, improving the safety of legged locomotion in microgravity. However, it does not suffice to guarantee stability during mobility. Therefore, we propose adding the Whole-Body Swing Momentum Distribution~(MD) technique to avoid pulling forces in the grippers, forming the RAMP framework.

To distribute the momentum from the swing motion and ensure zero velocity conditions for the supporting grippers, we divide the total momentum from~(\ref{eq:momentum}) into the momenta of the base $\boldsymbol{\mathcal{L}}_{\text{base}}$, supporting legs $\boldsymbol{\mathcal{L}}_{\text{sup}}$, and swing legs $\boldsymbol{\mathcal{L}}_{\text{swing}}$, as shown by~(\ref{eq:momentum_base+legs}), where $n_{\text{sup}}$ and $n_{\text{swing}}$ denote the number of supporting legs and the number of swing legs, respectively.

\begin{equation}
  \begin{aligned}
    \boldsymbol{\mathcal{L}} = & 
    \boldsymbol{\mathcal{L}}_{\text{base}}  + \boldsymbol{\mathcal{L}}_{\text{sup}}   + \boldsymbol{\mathcal{L}}_{\text{swing}}  \\
    = & 
    \boldsymbol{H}_{b} \dot{\boldsymbol{x}}_{b} + 
    \sum_{i=1}^{n_{\text{sup}}} \boldsymbol{H}_{bm,i} \dot{\boldsymbol{\phi}}_i + 
    \sum_{j=1}^{n_{\text{swing}}} \boldsymbol{H}_{bm,j} \dot{\boldsymbol{\phi}}_j
   \end{aligned} 
   \label{eq:momentum_base+legs}
\end{equation}

We combine~(\ref{eq:momentum_base+legs}) with the forward kinematics equation, as shown in~(\ref{eq:forward_kin}), to merge the momenta of the base and support legs into two terms as a function of the velocity of the base and the supporting grippers $\dot{\boldsymbol{x}}_{e,i}$. By forcing the zero velocity condition to the supporting grippers, we obtain the relation in~(\ref{eq:gripper_constrain}), where $\boldsymbol{J}_{b}$ indicates the Jacobian matrix of the base, and $\boldsymbol{J}_{mi}$ represents the Jacobian of the manipulator part. 

\begin{equation}
    \dot{\boldsymbol{x}}_{e,i} = 
    \boldsymbol{J}_{b} \dot{\boldsymbol{x}}_{b} +  \boldsymbol{J}_{mi} \dot{\boldsymbol{\phi}}_{i}
    \label{eq:forward_kin}
\end{equation}

\begin{equation}
    \boldsymbol{\mathcal{L}} = 
    \left( \boldsymbol{H}_{b} - \sum_{i=1}^{n_{\text{sup}}} \boldsymbol{H}_{bm,i} \boldsymbol{J}_{mi}^{+} \boldsymbol{J}_{b} \right) \dot{\boldsymbol{x}}_{b} + 
    \sum_{j=1}^{n_{\text{swing}}} \boldsymbol{H}_{bm,j} \dot{\boldsymbol{\phi}}_j 
    \label{eq:gripper_constrain}
\end{equation}

If we assume that there is no change in the momentum, we can compute the velocity of the base that compensates for a fraction $\alpha$ of the momentum generated from the swing motion, as shown in~(\ref{eq:momentum_distribution}). In summary, we distribute the swing momentum to the base link and supporting legs, avoiding that high motion reactions cause the detachment of the gripper by securing zero velocity for supporting end-effectors.

\begin{equation}
    \left( \boldsymbol{H}_{b} - \sum_{i=1}^{n_{\text{sup}}} \boldsymbol{H}_{bm,i} \boldsymbol{J}_{mi}^{+} \boldsymbol{J}_{b} \right) \dot{\boldsymbol{x}}_{b} = 
    -\alpha \sum_{i=1}^{n_{\text{swing}}} \boldsymbol{H}_{bm,j} \dot{\boldsymbol{\phi}}_j
    \label{eq:momentum_distribution}
\end{equation}

If the momentum distribution factor $\alpha$ is equal to zero in~(\ref{eq:momentum_distribution}), we have a base velocity equal to zero, which is the same as not performing any momentum distribution. By selecting an appropriate value for the momentum distribution factor greater than 0 and smaller than 1, it is possible to reduce a significant amount of motion reactions in the gripper, keeping the base attitude within safe limits to avoid singularities. We name this procedure Reaction-Aware Motion Planning with Partial Momentum Distribution~(RAMP-PMD). If $\alpha$ equals 1, the swing momentum is fully distributed to the remaining parts of the robot, mitigating the motion reactions. We call this technique RAMP with Full Momentum Distribution~(RAMP-FMD).

With RAMP, we calculate the position for the swing end-effectors using LRST in~(\ref{eq:opt_bez}). Moreover, we compute the necessary base velocity in~(\ref{eq:momentum_distribution}) to distribute the swing reactions based on the momentum calculated from the moving legs' speed. Therefore, considering the zero velocity condition for the supporting grippers, RAMP outputs the trajectory for all end-effectors and the base, considerably mitigating the detachment risk for mobility in microgravity.

\section{Simulation Studies}

Testing new technologies and strategies for locomotion in microgravity is extremely difficult due to the limitations in sending equipment for tests in space. Therefore, numerical simulations are essential in developing new mobility approaches for robots to explore SSSBs and assist in space station activities.

In this paper, we use ClimbLab as a simulation platform. ClimbLab is a dynamic simulator developed in MATLAB for climbing-legged robots~\cite{uno2021climblab}. ClimbLab uses SpaceDyn to compute the dynamics of multi-body robots and other parameters, such as inertia and Jacobian matrices~\cite{yoshida1999spacedyn}.

The main criterion for failure analysis in the simulations is the detachment of the gripper due to excessive pulling forces. In this model, if a ground reaction force acting on the robot pulls it towards the surface, exceeding the maximum holding force of the gripper, a detachment occurs. We use a compliant contact model to compute ground reaction forces, assuming the end-effector as a single point and the initial contact point as the neutral position.

\subsection{Quadrupedal Robot on an Asteroid (Rough Surface)}

For this simulation study, we assume a rough surface to emulate an asteroid with a gravity of $10^{-6}$~G. The surface coefficients for stiffness and damping are 4000~N/m and 1~Ns/m, respectively. As for the robot model, we assume the inertial and length parameters of Hubrobo, a quadruped insect-type robot with three controllable articulations per leg~\cite{uno2021hubrobo}. Details of the parameters of the robot are in Table~\ref{tab:hubrobo}. For this simulation, we assume a considerably small holding force of 0.9~N, as the fragile and porous surface of an SSSB requires weak grasping.

\begin{table}[b]
\caption{Inertial Parameters of Quadrupedal Robot}
\label{tab:hubrobo}
\begin{center}
\begin{tabular}{c c c c c c}
Link & Size (mm) & Mass (g) & &  Inertia (kgm$^2$) & \\
     &           &          & $I_{xx}$ & $I_{yy}$ & $I_{zz}$\\
\hline
Base & 108 x 108   & 441.7    & 9.2e-4 & 1.4e-3 & 1.3e-3\\
1 & 28.5 x 17.5 & 205.6   & 4.2e-5 & 2.8e-5 & 2.8e-5\\
2 & 107 & 27.3          & 1.5e-5 & 2.4e-5 & 3.7e-5\\
3 & 143 & 220.2         & 2.2e-5 & 1.8e-5 & 1.2e-5\\
\end{tabular}
\end{center}
\end{table}

The robot walks with a periodic crawl gait, moving first the hind legs and then the front legs, with a fixed stride of 8~cm and a step height of 4~cm. For this study, we decouple the onward motion of the base and the swing leg so the RAMP can distribute the swing momentum to the whole body, including the base link. Therefore, between swing movements, we have a phase where the base moves 2~cm from its previous neutral position, adjusting the position and attitude to complete a periodic cycle after the motion of the four legs. We also implemented release and grasping vertical movements of 1~cm each to avoid ground collisions with the swing trajectory. The total swing period of each leg is 1.5~s, the same as the period of each base movement, adding to 12~s for one cycle period of this quadruped robot.

\begin{figure}[tb]
\centering
\subfigure
{\includegraphics[width=0.98\linewidth]{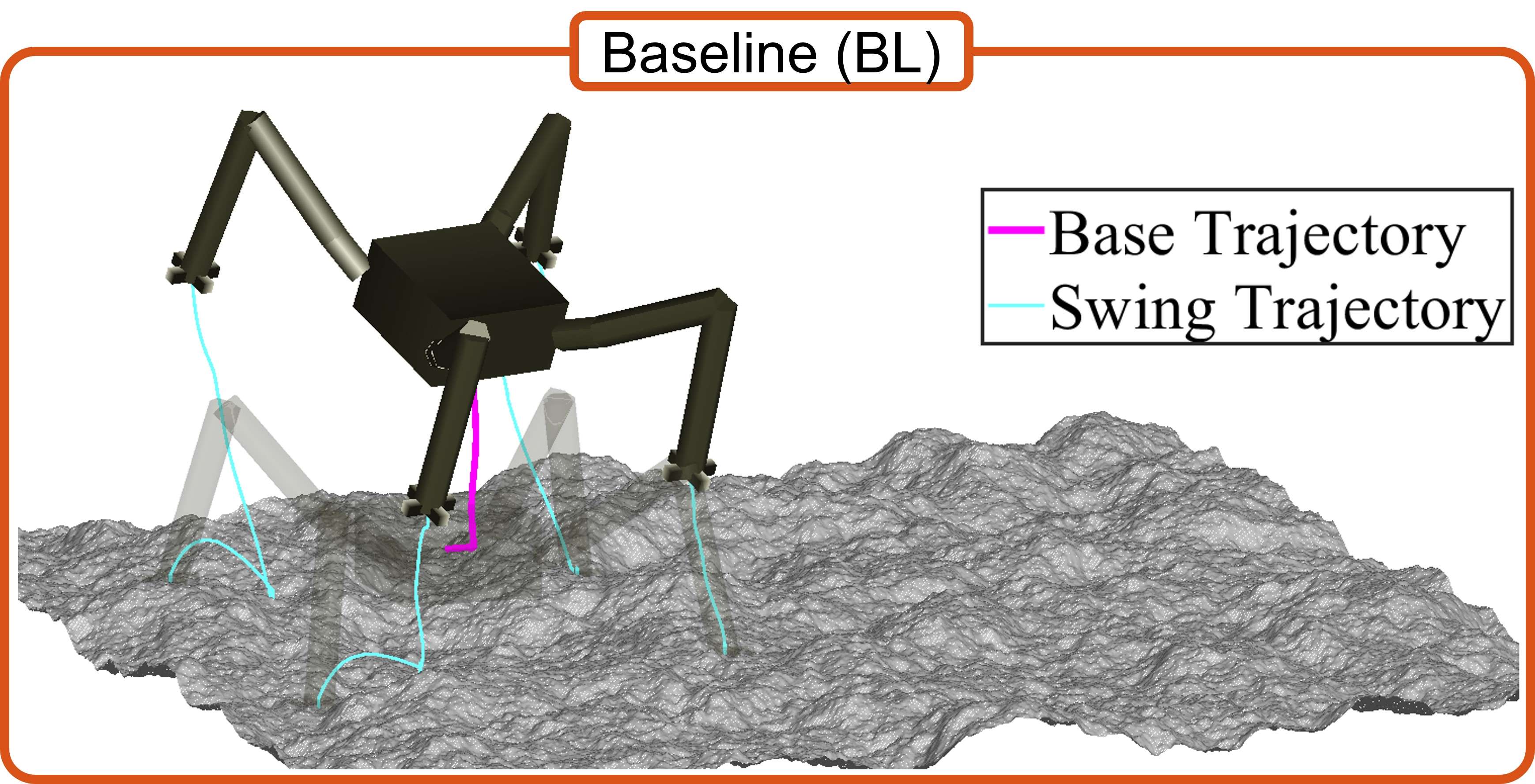}
\label{fig:BL_sim}
}\\
\vspace{-3mm}
\subfigure
{\includegraphics[width=0.98\linewidth]{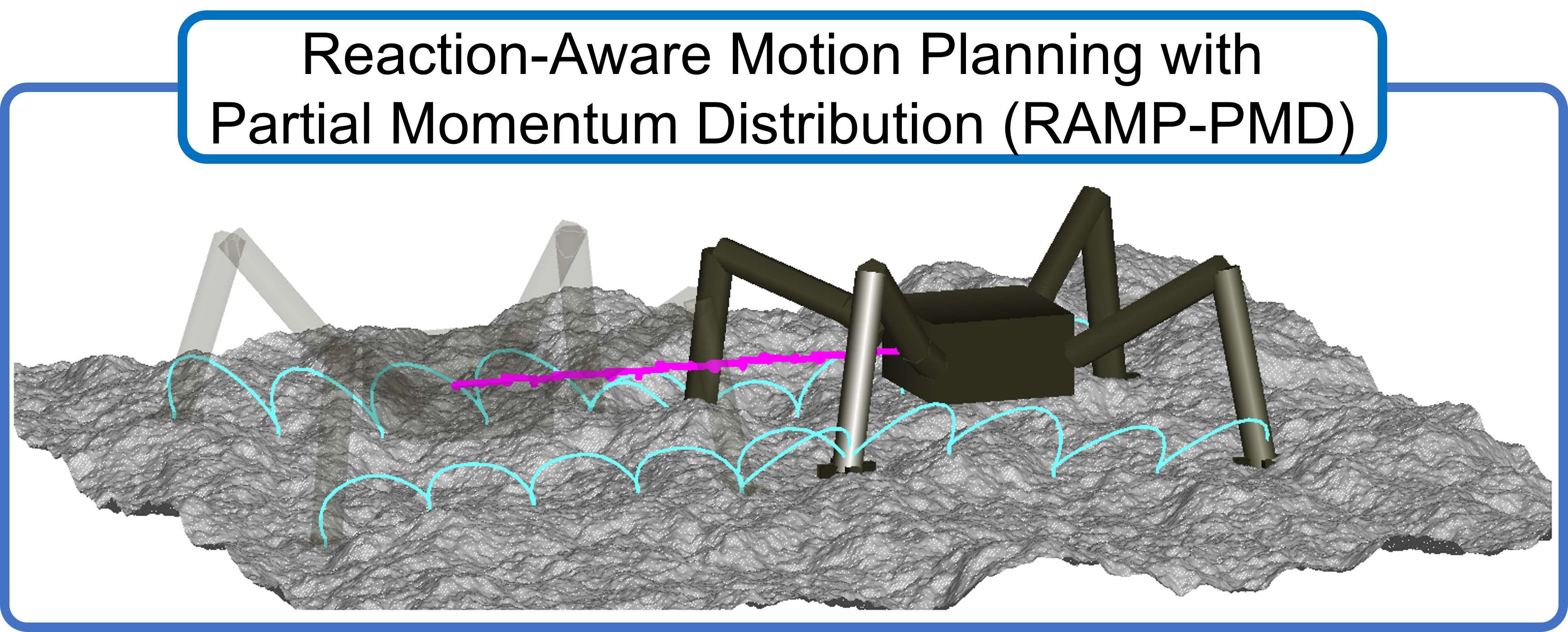}
\label{fig:PMD_sim}
}\\
\vspace{-3mm}
\subfigure
{\includegraphics[width=0.98\linewidth]{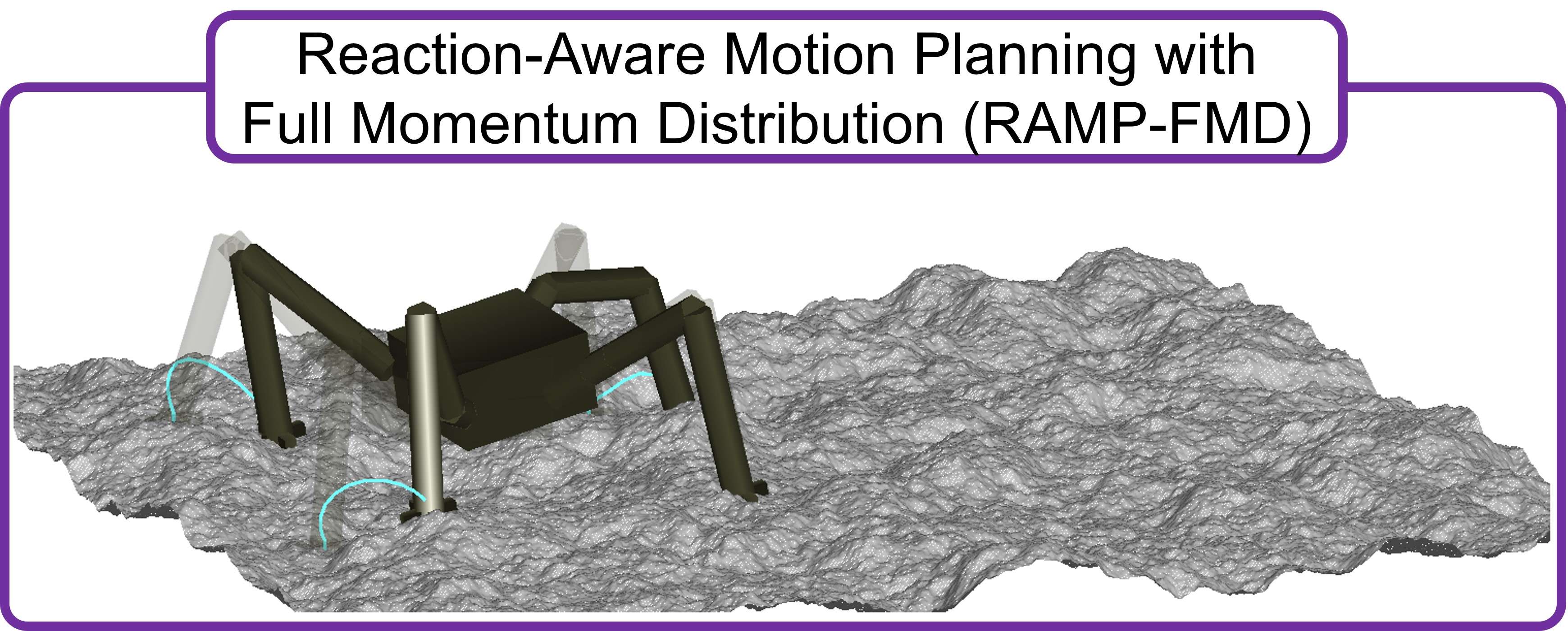}
\label{fig:FMD_sim}
}
\caption{Snapshots of the simulation study for rough surface locomotion with a quadruped robot. Blue lines represent the swing trajectory, and pink is the base trajectory. The baseline (BL) case fails due to detachment, inducing flotation on the robot, while the proposed method RAMP with Partial Momentum Distribution (RAMP-PMD) successfully reaches the goal position. RAMP with Full Momentum Distribution (RAMP-FMD) fails due to a singularity.}
\label{fig:sim}
\end{figure}

Fig.~\ref{fig:sim} shows snapshots of three different cases of simulations. The first is a baseline (BL) case without any RAMP element, simply computing swing trajectories from the current to the next grasping position through a via point at the desired height. The second case performs the Reaction-Aware Motion Planning with a Partial Momentum Distribution (RAMP-PMD) with a distribution factor of 0.5. And the last case is also a RAMP, but with Full Momentum Distribution (RAMP-FMD), i.e., the distribution factor is equal to 1. After RAMP computes the desired trajectories for the robot, an inverse kinematics algorithm computes the angular positions of each joint, which are the inputs of a PD torque controller to execute the motion.

The results show that after a few steps, walking with the regular baseline motion plan leads to ground reaction forces higher than the gripper holding limit, causing detachment and flotation of the robot. This failure case is highly undesired in real scenarios, as this could lead to a complete loss of the robot during the exploration mission. By planning the motion with RAMP-PMD, the robot successfully advances 40 cm without exceeding force limits. However, in the last case with RAMP-FMD, the robot moves its base excessively to distribute the entire swing momentum, falling into a singularity configuration.

The simulation results show that RAMP can effectively increase the safety of legged mobility in microgravity, transforming a failure case into a successful walk until the desired location. However, the distribution factor has to be tuned appropriately to avoid a swing momentum distribution beyond the robot's capabilities, considering parameters such as inertia, moving speed, and manipulability.

\section{Experimental Verification}

Although it is difficult to emulate microgravity conditions accurately on Earth, it is possible to create a planar environment by eliminating the friction between a robot and an extremely flat surface. A common technique is to levitate the robot with pressurized air, avoiding direct contact with the flat surface. We use such system to perform two-dimensional verification of the RAMP method to improve mobility in microgravity.

\begin{figure}[t]
\centering
\includegraphics[width=0.48\textwidth]{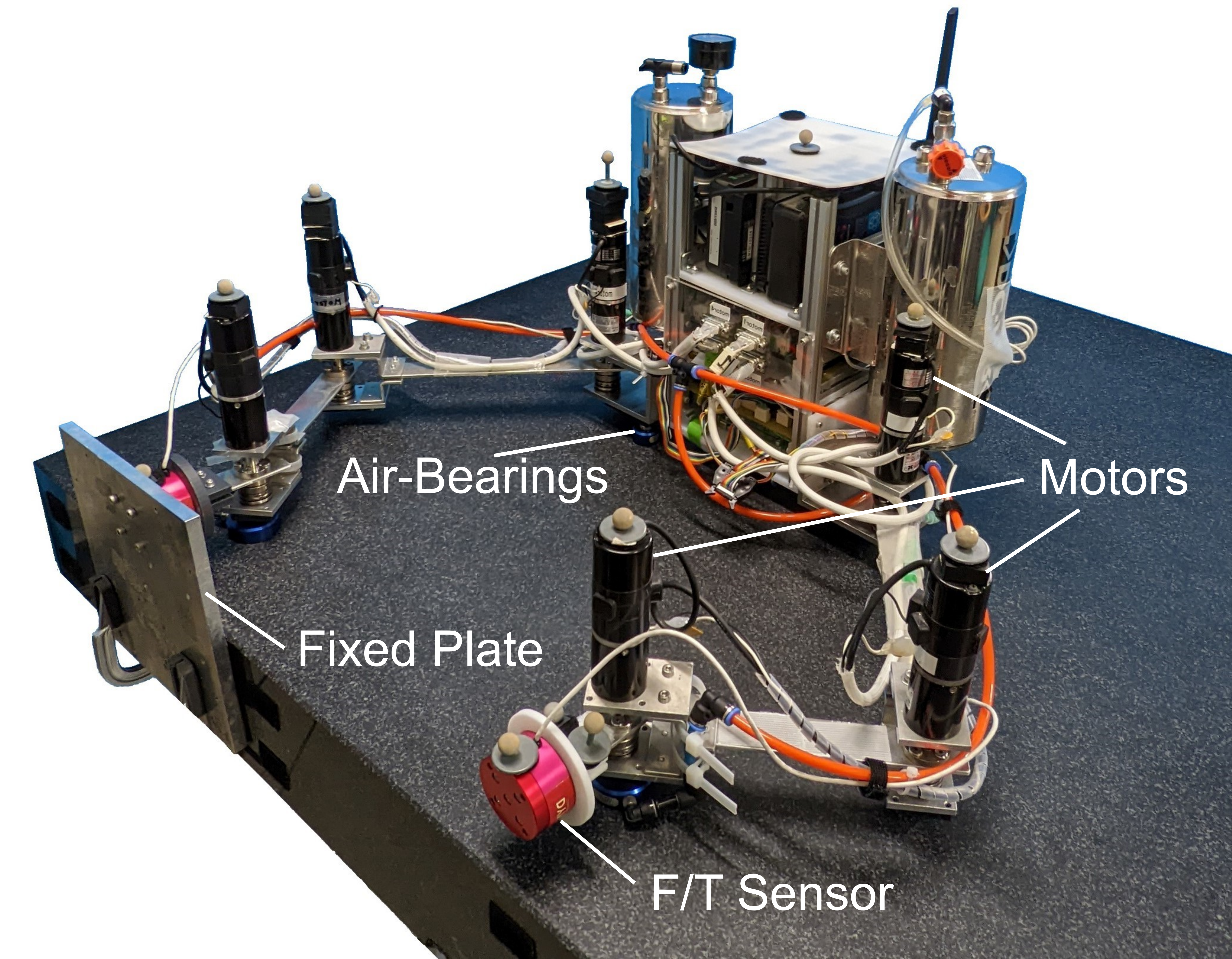}
\caption{Dual-Arm Air-Floating Robot. The robot floats with pressurized air from the air-bearings, eliminating friction to create a two-dimensional microgravity environment. Each limb has 3 degrees of freedom controllable by DC motors, and a force/torque sensor at its tip. One of the limbs is directly attached to a fixed plate to emulate the grasping of a surface.}
\label{fig:DAR}
\end{figure}

\subsection{Planar Robot with Fixed Leg}

\begin{table}[b]
\caption{Inertial Parameters of Dual-Arm Air-Floating Robot}
\label{tab:DAR}
\begin{center}
\begin{tabular}{c c c c}
Link & Size (mm) & Mass (g) &   Inertia (kgm$^2$)  \\
     &           &          & $I_{zz}$ \\
\hline
Base & 160 x 295     & 8524.0    & 1.2e-1 \\
1 & 25          & 608.0   & 1.6e-3 \\
2 & 17.5        & 626.2   & 7.1e-4 \\
3 & 87.25       & 248.1   & 2.0e-4 \\
\end{tabular}
\end{center}
\end{table}

For this experimental study, we use the Dual-Arm Air-Floating Robot shown in Fig.~\ref{fig:DAR}, with size and inertial parameters detailed in Table~\ref{tab:DAR}. The robot has three Degrees of Freedom per limb, with a force and torque (F/T) sensor at its tip. We attach one of the limbs to a fixed plate to emulate grasping without detachment to measure ground reactions. Since the robot cannot walk continuously, we only perform a single-step analysis. The step has a 15~cm stride and 7~cm height, with a 2~s swing period. We compute the trajectories of the base and the limbs, and the inverse kinematics offline, while the robot applies a velocity control based on the desired joint positions.

\begin{figure}[tb]
\centering
\subfigure
{\includegraphics[width=0.98\linewidth]{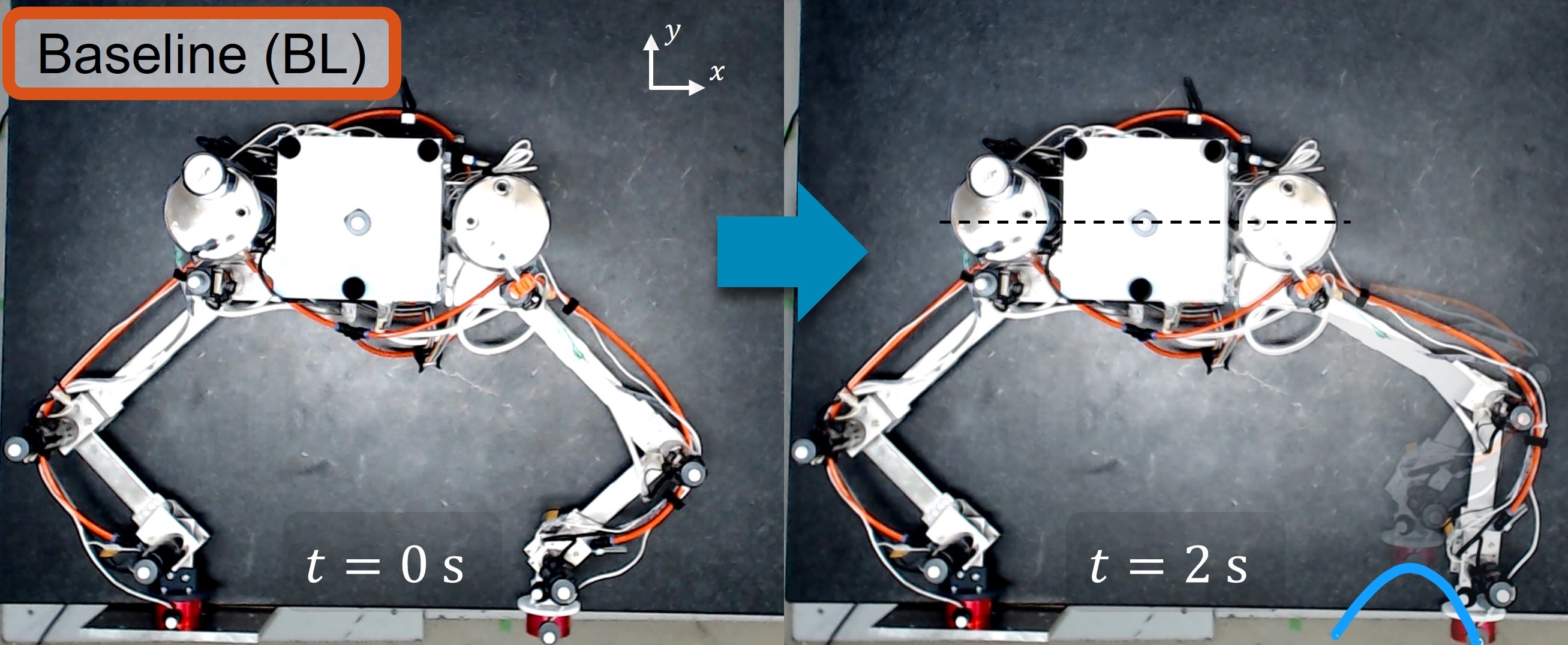}
\label{fig:BL}
}\\
\vspace{-3mm}
\subfigure
{\includegraphics[width=0.98\linewidth]{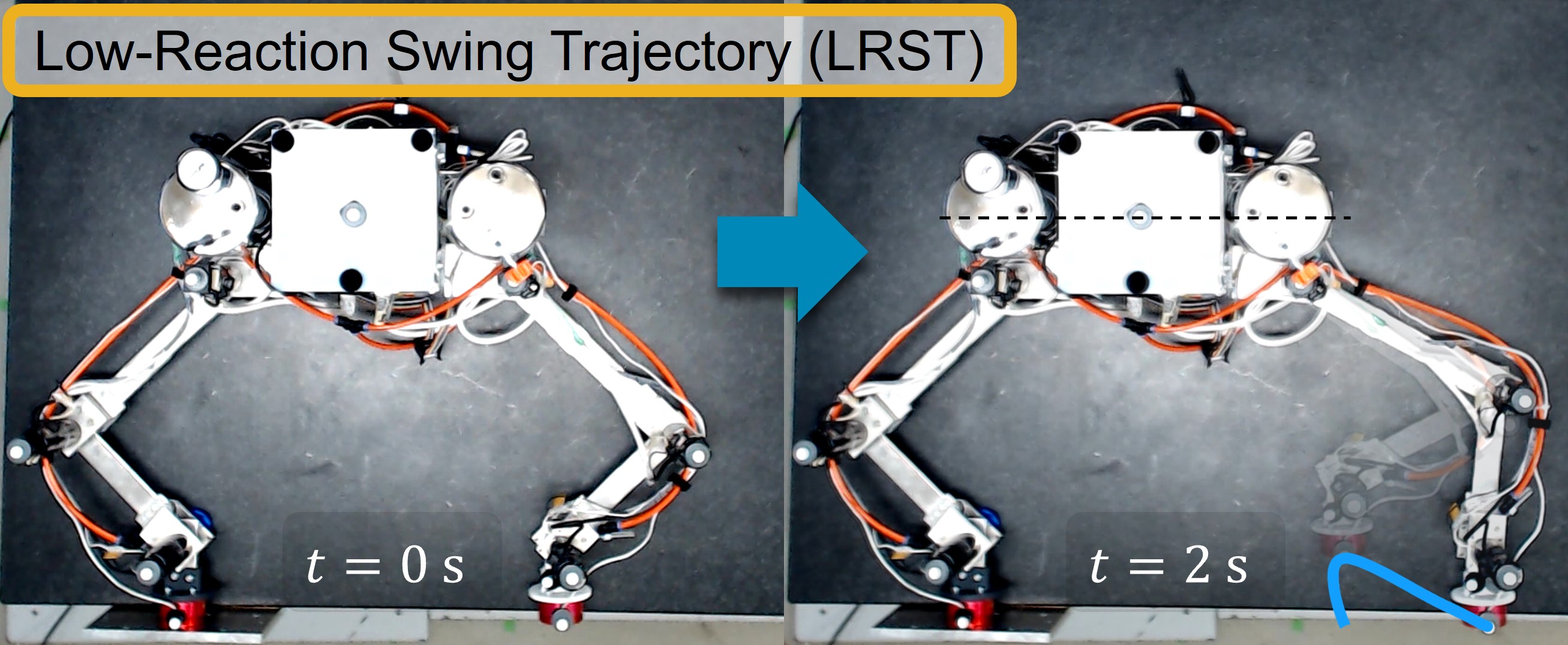}
\label{fig:LRST}
}\\
\vspace{-3mm}
\subfigure
{\includegraphics[width=0.98\linewidth]{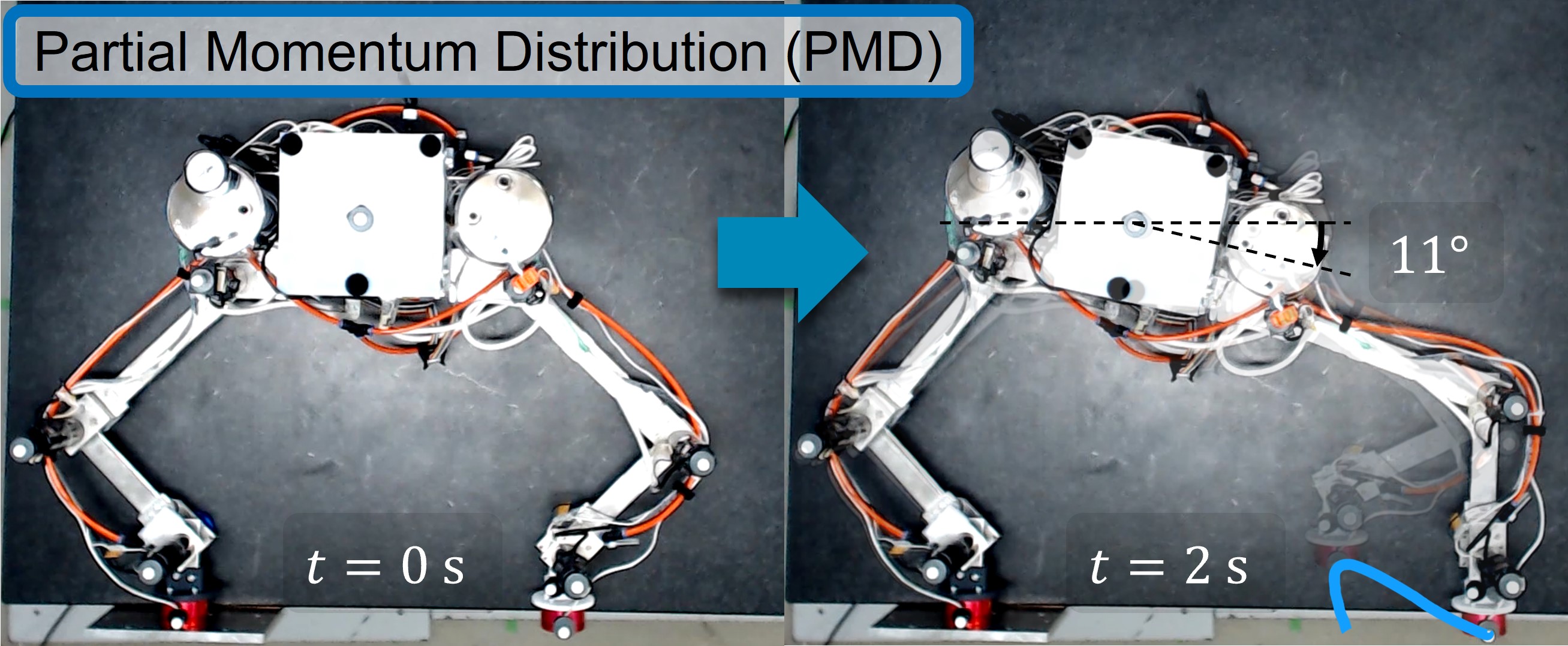}
\label{fig:PMD}
}\\
\vspace{-3mm}
\subfigure
{\includegraphics[width=0.98\linewidth]{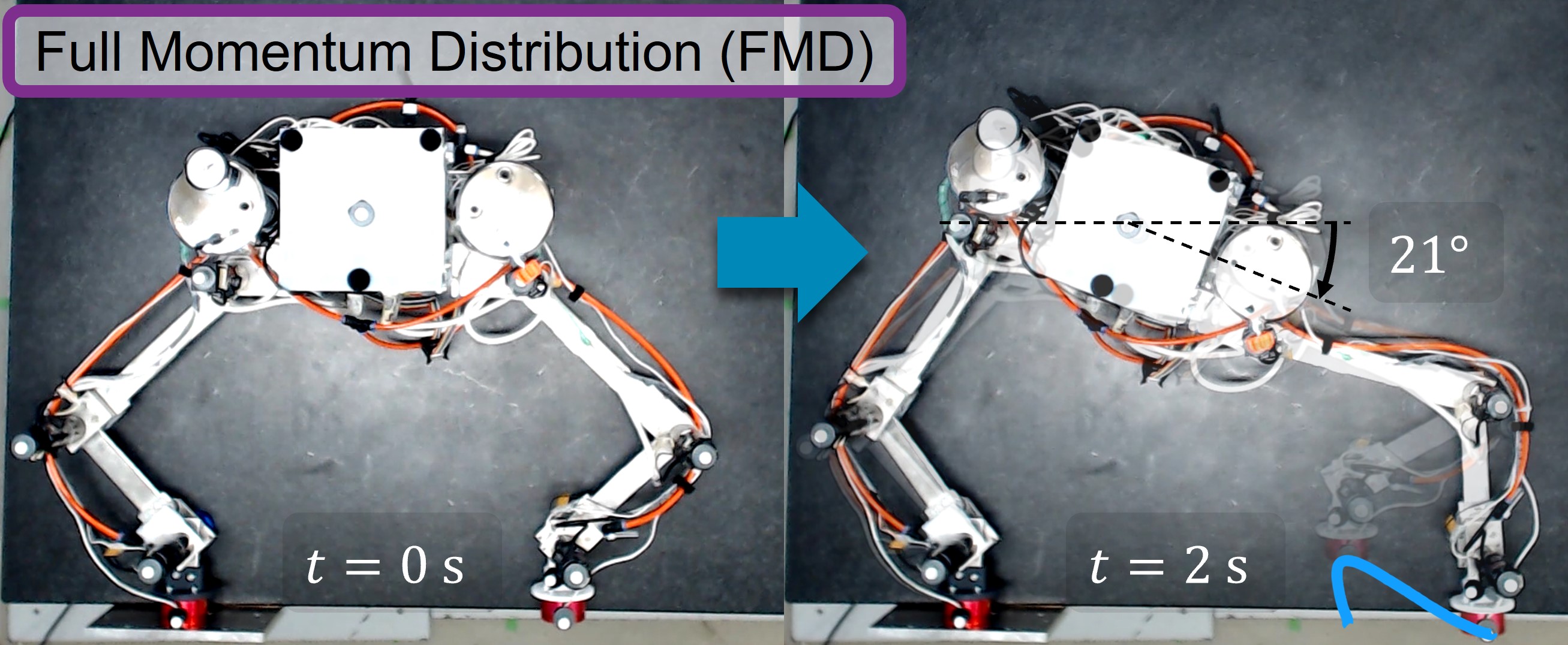}
\label{fig:FMD}
}
\caption{Snapshots of all experiments cases using the Dual-Arm Air-Floating Robot. The left limb is fixed directly to a plate, and the right leg moves along the planned trajectory (blue line) during 2 s. The baseline (BL) case moves the leg with a standard spline trajectory, while the Low-Reaction Swing Trajectory (LRST) case moves the swing limb in a trajectory that minimizes the momentum variation. The partial Momentum Distribution (PMD) case also moves the base and support limb to distribute half of the swing momentum to the whole body. The Full Momentum Distribution (FMD) case distributes 100\% of the swing momentum.}
\label{fig:exp}
\end{figure}

Fig.~\ref{fig:exp} shows snapshots of four distinct experiment cases. The first is the Baseline (BL) case with a standard trajectory and no momentum distribution, while the other three use RAMP with different momentum distribution factors. The second case is the Low-Reaction Swing Trajectory (LRST), which has only the optimized trajectory but no momentum distribution. The third and fourth cases are Partial Momentum Distribution (PMD) and Full Momentum Distribution (FMD), with momentum distribution factor of 0.5 and 1, respectively.

We can observe the differences between low-reaction and standard trajectories by comparing the four cases. The LRST minimizes the motion reactions by reducing the limb's inertia when bringing it closer to the main body. Moreover, we notice how the robot moves its whole body to distribute the swing momentum in the last two cases, showing an attitude change of $11^\circ$ and $21^\circ$ for PMD and FMD, respectively. As this robot has higher inertia and manipulability when compared to the robot used in the simulation study, we do not observe any problem related to singular configurations, even with the robot moving at high speeds.

\begin{figure}[tb]
\centering
\subfigure
{\includegraphics[width=0.48\linewidth]{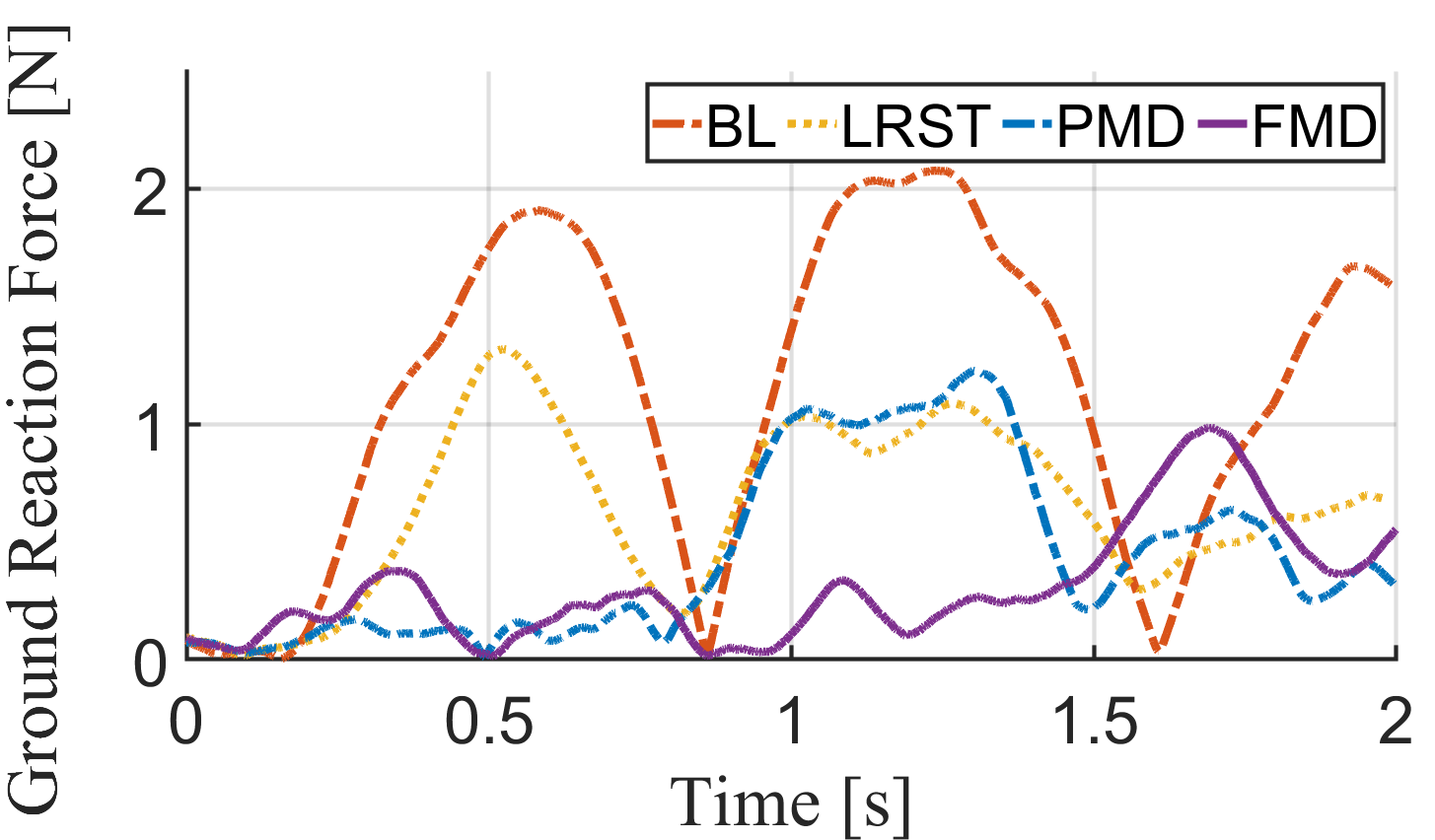}
\label{fig:F_res}
}
\hspace{-2mm}
\subfigure
{\includegraphics[width=0.48\linewidth]{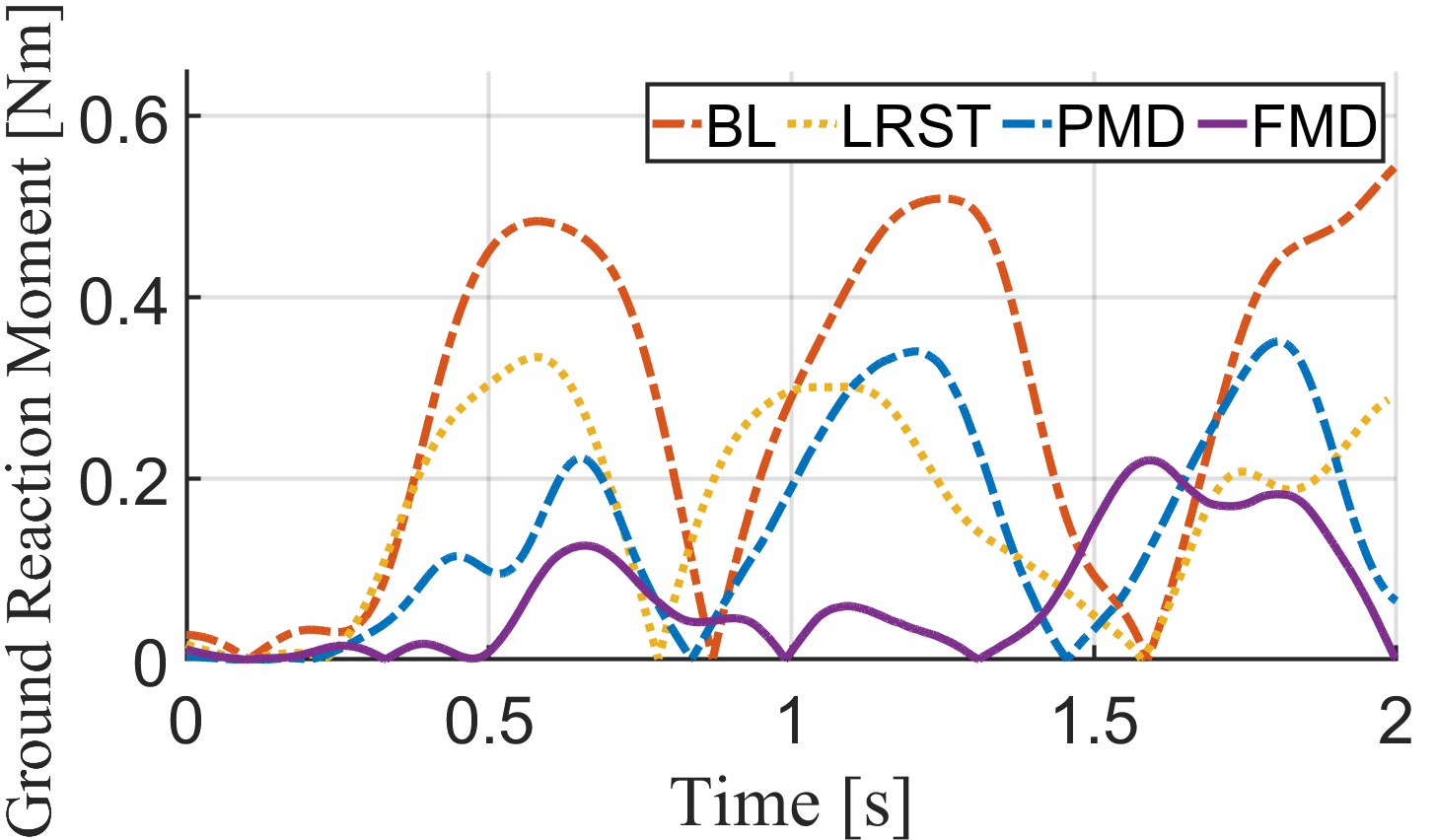}
\label{fig:T_res}
}
\caption{Time-history graphs of Ground Reaction Force and Moment of the experimental study. Baseline (BL) is in red, Low-Reaction Swing Trajectory (LRST) is in yellow, Partial Momentum Distribution (PDM) is in blue, and Full Momentum Distribution (FMD) is in purple. We verify that by increasing the degree of the Reaction-Aware Motion Planning (RAMP) with low-reaction swing trajectory and momentum distribution, the reaction forces and moments acting on the support limb are mitigated.}
\label{fig:FT_res}
\end{figure}

\begin{figure}[tb]
\centering
\subfigure
{\includegraphics[width=0.48\linewidth]{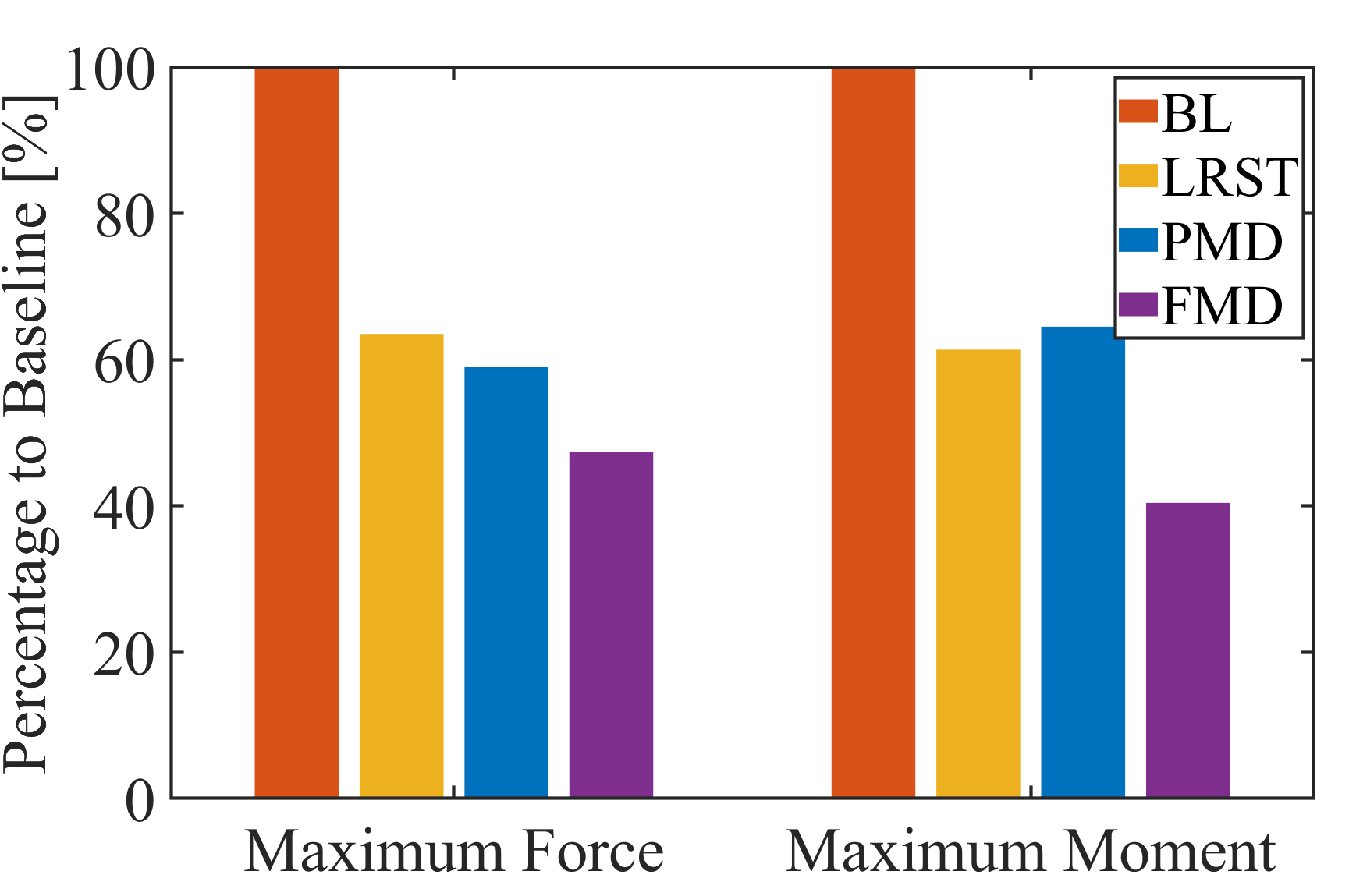}
\label{fig:FT_ave}
}
\hspace{-2mm}
\subfigure
{\includegraphics[width=0.48\linewidth]{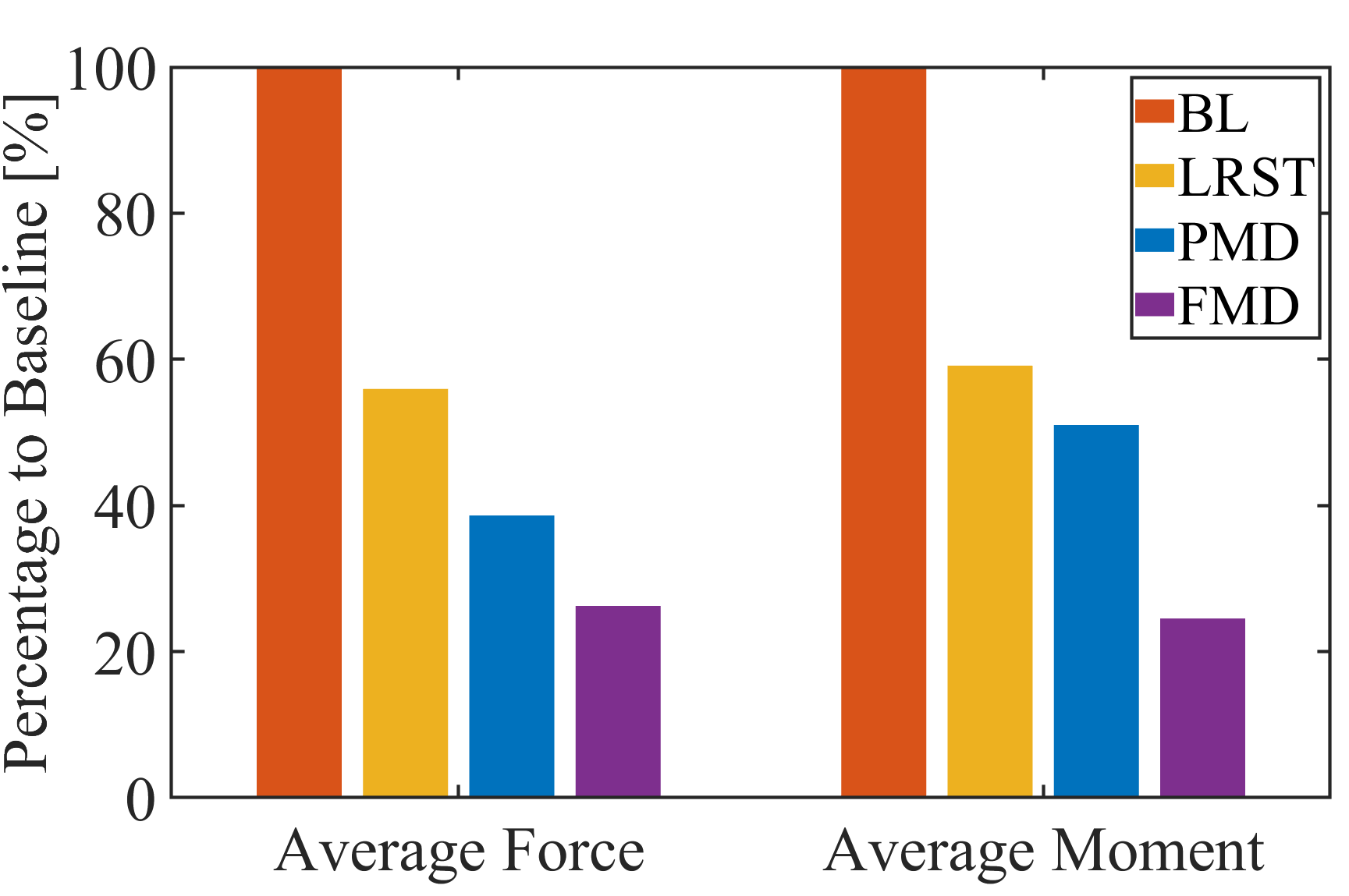}
\label{fig:FT_max}
}
\caption{Summary of maximum and average ground reactions in relation to the Baseline case. Both forces and moments are reduced by employing RAMP, with a progressive decrease of maximum and average values as we increase the momentum distribution factor. With FMD, it is possible to decrease up to 60\% of the maximum ground reactions and up to 75\% of average ground reactions.}
\label{fig:FT_ave/max}
\end{figure}

Fig.~\ref{fig:FT_res} and Fig.~\ref{fig:FT_ave/max} show the data obtained from the F/T sensor of the fixed limb. Compared to the baseline case shown in red, RAMP-FMD can reduce the maximum ground reactions to 40\% of the force and torque observed in the Baseline case. RAMP also considerably decreases average force and moment, ranging from 60\% to 25\% of the baseline average, depending on the momentum distribution factor.

RAMP demonstrated its efficacy in decreasing ground reaction forces and moments with the experimental results of a planar robot in an emulated microgravity. From these results, we can expect that RAMP can effectively increase the locomotion safety of multi-legged robots in a real scenario by reducing the detachment risk of the gripper due to high pulling forces.

\section{Conclusions}

We propose a novel Reaction-Aware Motion Planning (RAMP) for the mobility of multi-legged robots in microgravity, computing the optimal trajectories for the swing legs and base that mitigate motion reactions. The main contribution of our approach is improving the safety of dynamic multi-legged locomotion in microgravity environments by decreasing the risk of surface separation. RAMP also provides low-reaction alternatives to reactionless control strategies with an adjustable momentum distribution factor to avoid singular configurations. We demonstrated the efficacy of our approach in preventing surface detachment through numerical simulations and experiments, indicating that real robotic mission scenarios can employ RAMP for safe mobility in microgravity.

\addtolength{\textheight}{-18cm}   



\bibliography{./IEEEabrv,reference.bib}

\end{document}